\documentclass[conference]{IEEEtran}
\IEEEoverridecommandlockouts
\usepackage{cite}
\usepackage{amsmath,amssymb,amsfonts}
\usepackage{algorithmic}
\usepackage{graphicx}
\usepackage{textcomp}

\DeclareMathOperator*{\argmin}{arg\,min}

\usepackage{xcolor}
\usepackage[pagebackref=true,breaklinks=true,colorlinks,bookmarks=false]{hyperref}
\begin{document}

\title{Resource Efficient Mountainous Skyline Extraction using Shallow Learning}
%\title{Mountainous Skyline Extraction using Structure Tensor based Learnable Filters}
%\title{Mountainous Skyline Extraction using Learnable Filters}
%\title{Mountainous Skyline Extraction using Shallow Learning}

\makeatletter
\newcommand{\linebreakand}{%
  \end{@IEEEauthorhalign}
  \hfill\mbox{}\par
  \mbox{}\hfill\begin{@IEEEauthorhalign}
}
\makeatother

%\author{\IEEEauthorblockN{Touqeer Ahmad}
%\IEEEauthorblockA{\textit{Department of Computer Science}\\
%\textit{University of Colorado at Colorado Springs}\\
%Colorado Springs, USA\\
%touqeer@vast.uccs.edu}
%\and
%\IEEEauthorblockN{Ebrahim Emami}
%\IEEEauthorblockA{\textit{Department of Computer Science and Engineering}\\
%\textit{University of Nevada, Reno}\\
%Reno, USA\\
%ebrahim@nevada.unr.edu}
%\linebreakand
%\and
%\IEEEauthorblockN{Martin Čadík}
%\IEEEauthorblockA{\textit{Faculty of Information Technology}\\
%\textit{Brno University of Technology}\\
%Brno, Czech Republic\\
%cadik@fit.vutbr.cz}
%\and
%\IEEEauthorblockN{George Bebis}
%\IEEEauthorblockA{\textit{Department of Computer Science and Engineering}\\
%\textit{University of Nevada, Reno}\\
%Reno, USA\\
%bebis@cse.unr.edu}
%}

\author{\IEEEauthorblockN{Touqeer Ahmad\IEEEauthorrefmark{1}, Ebrahim Emami\IEEEauthorrefmark{2}, Martin Čadík\IEEEauthorrefmark{3}, George Bebis\IEEEauthorrefmark{2}}
\IEEEauthorblockA{\IEEEauthorrefmark{1}Vision and Security Technology Lab, University of Colorado at Colorado Springs, USA}
touqeer@vast.uccs.edu
\IEEEauthorblockA{\IEEEauthorrefmark{2}Department of Computer Science and Engineering, University of Nevada, Reno, USA}
ebrahim@nevada.unr.edu, bebis@cse.unr.edu
\IEEEauthorblockA{\IEEEauthorrefmark{3} Faculty of Information Technology, Brno University of Technology, Czech Republic}
cadik@fit.vutbr.cz
}

\maketitle

\begin{abstract}
Skyline plays a pivotal role in mountainous visual geo-localization and localization/navigation of planetary rovers/UAVs and virtual/augmented reality applications. We present a novel mountainous skyline detection approach where we adapt a shallow learning approach to learn a set of filters to discriminate between edges belonging to sky-mountain boundary and others coming from different regions. Unlike earlier approaches, which either rely on extraction of explicit feature descriptors and their classification, or fine-tuning general scene parsing deep networks for sky segmentation, our approach learns linear filters based on local structure analysis. At test time, for every candidate edge pixel, a single filter is chosen from the set of learned filters based on pixel's structure tensor, and then applied to the patch around it. We then employ dynamic programming to solve the shortest path problem for the resultant multistage graph to get the sky-mountain boundary. The proposed approach is computationally faster than earlier methods while providing comparable performance and is more suitable for resource constrained platforms e.g., mobile devices, planetary rovers and UAVs.  We compare our proposed approach against earlier skyline detection methods using four different data sets. Our code is available at \url{https://github.com/TouqeerAhmad/skyline_detection}. 
\end{abstract}

%\begin{IEEEkeywords}
%component, formatting, style, styling, insert
%\end{IEEEkeywords}

%\crushmath
\section{Introduction}
Skyline detection is the problem of finding a path that extends from left-most column of an image to the right-most and divides an image into sky and non-sky regions. Finding skyline is a challenging vision problem due to non-linear nature of the skylines, variations in the sky and non-sky regions due to geographical terrains and weather conditions. Skyline detection serves as the underlying method for many practical applications and have been investigated for navigation and localization of Unmanned  Aerial Vehicles (UAVs) \cite{Boroujeni12,McGee05,Thurrowgood09,Grelsson15,Hou15,Di12,Croon11,Ettinger02,Todorovic03}, planetary rover and vehicle localization \cite{Boukas14,Cozman97,Cozman00,Gupta08,Ho14,Dumble12,Dumble15,Neto11}, augmented reality and tourism applications \cite{Baboud11,Porzi14,Brejcha18}, geolocation of mountain and desert images \cite{Chen15,Saurer2016,Baatz2012,Liu14,Tzeng13,Baboud11}, marine security and ship detection \cite{Fefilatyev06,Gershikov13,Kruger10,Kong16}. 

Sky segmentation and skyline detection are two related yet distinct problems. In a general scene parsing sense, sky segmentation is equivalent to semantically identifying all the pixels belonging to the sky region where the sky region may or may not extend from left-to-right and may be present in a small portion of the image. In the case of the skyline extraction problem, a skyline is thought of as a \textbf{visible} boundary which extends from \textbf{left-to-right} and divides the input image into two main regions i.e. sky and non-sky. Depending upon the specificity of the non-sky region, the skyline could refer to the mountainous skyline, sea-shore skyline or city skyline. While any general scene-parsing network can be fine-tuned for sky segmentation as demonstrated in \cite{Ahmad2017}, the two problems are not inter-changeable, unless the assumptions for skyline hold true.     

Recent efforts for skyline detection can be grouped into two main categories. The first group of approaches employs supervised learning to discriminate between skyline and non-skyline pixels by either using feature descriptors or directly discriminating based on the pixel intensities \cite{Ahmad2013,Ahmad2014,Ahmad2015,Ahmad2015b,Ahmad2015c,Ahmad2020,Hung13}. Some approaches belonging to this category use edge information while others solely rely on discrimination power of the trained classifier. These approaches generally address skyline detection as a shortest-path problem and incorporate dynamic programming (DP) which was first utilized for skyline detection problem in \cite{Lie05}. The second type of approaches address the skyline detection in the context of sky-segmentation problem and generally rely on Convolutional Neural Networks (CNNs) to segment out the sky regions \cite{Porzi14,Porzi16,Verbickas14}. There have also been some attempts to fine-tune semantic segmentation networks for sky segmentation \cite{Ahmad2017}. While both types of methods perform reasonably well for skyline detection, they are computationally expensive solutions e.g., in some cases, single or multiple feature descriptors around each pixel need to be extracted before passing them through the trained classifier (Support Vector Machines/Convolutional Neural Networks) or directly employing dense deep networks which are inherently quite expensive. Due to expensive computational and memory requirements, such approaches are not very suitable for resource-constrained platforms e.g., mobile devices, UAVs and planetary rovers. To address these issues, in this paper we present a skyline detection approach which builds upon a recently introduced \textbf{shallow learning} framework \cite{Getreuer2018} targeted for \textbf{resource-constrained} devices and provides a reasonable trade-off between computational cost and performance.\par     

The rest of the paper is structured as follows: Section \ref{sec_2} provides the details of earlier work on skyline detection methods relying on dynamic programming (DP) and establishes the foundation for our proposed approach and its comparisons. In Section \ref{sec_3}, we describe the shallow learning framework employed for our proposed skyline detection approach. Experimental details and results are provided in Section \ref{sec_4}. The paper is concluded with pointers for future work in Section \ref{sec_5}.

%-------------------------------------------------------------------------

\begin{figure*}[t]
\begin{center}
\includegraphics[width=2.26in]{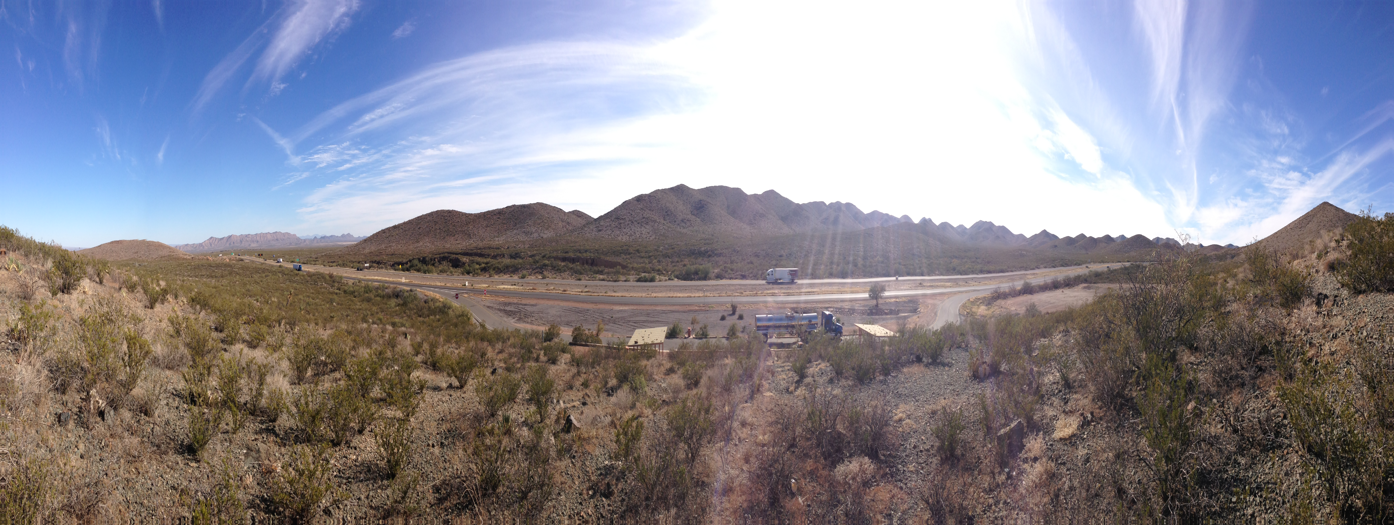}
\includegraphics[width=2.26in]{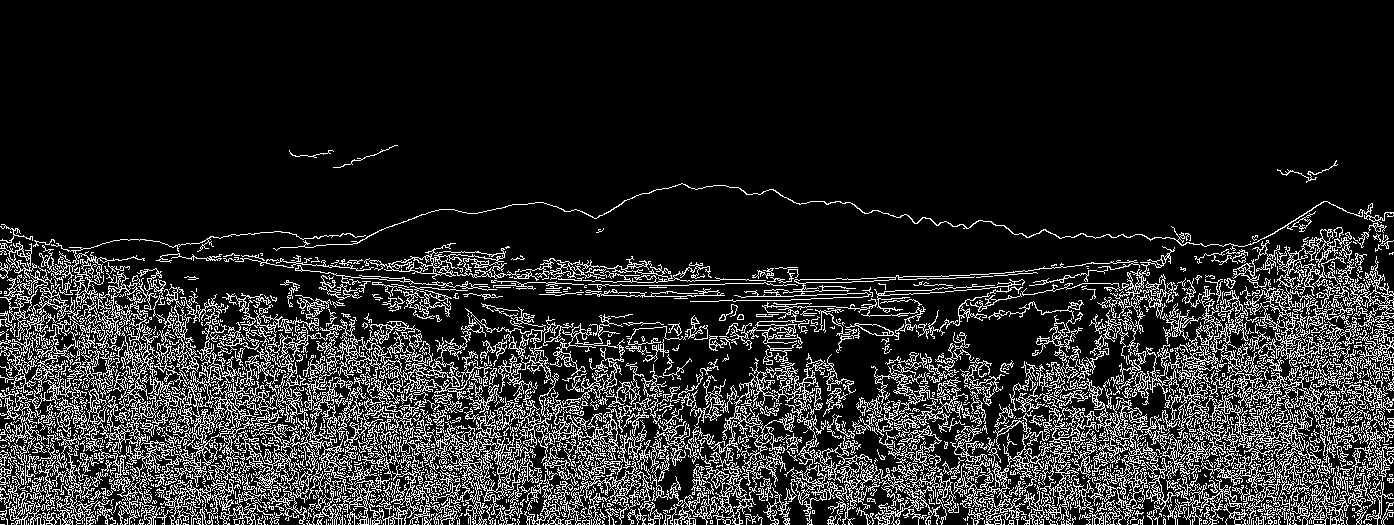}
\includegraphics[width=2.26in]{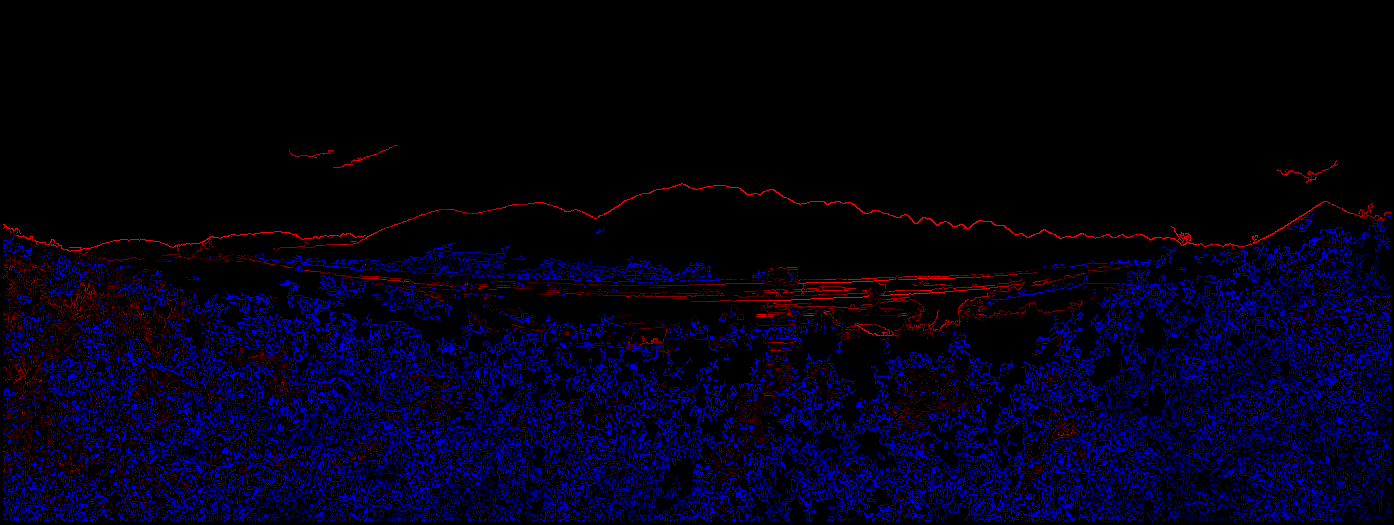}\\
(a) \hspace{2.26in} (b) \hspace{2.26in} (c)\\
\includegraphics[width=2.26in]{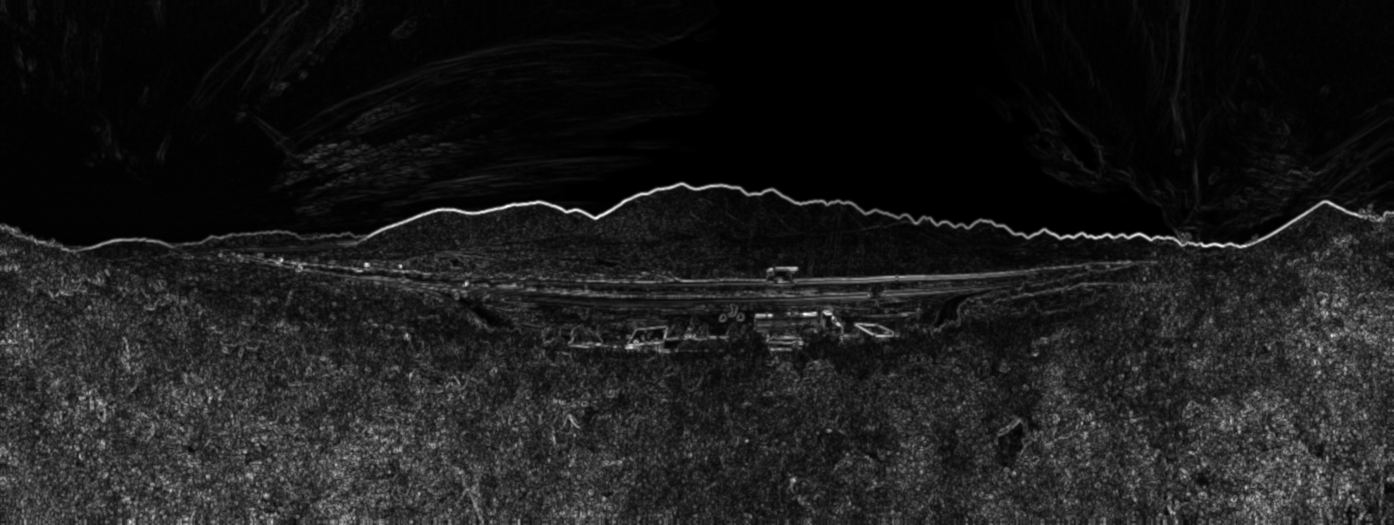}
\includegraphics[width=2.26in]{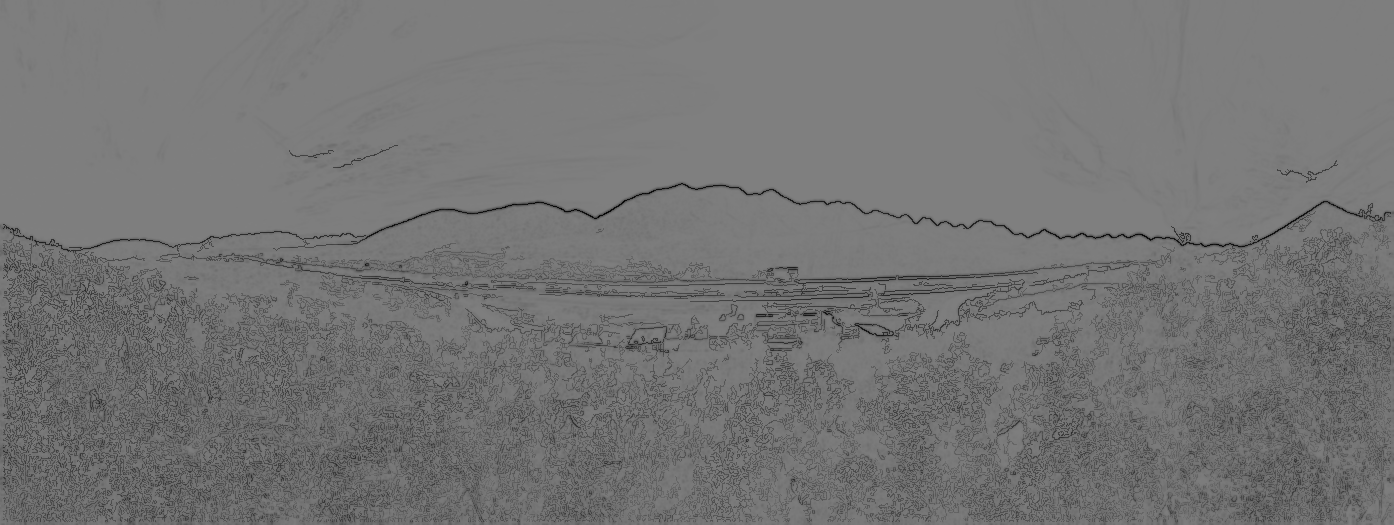}
\includegraphics[width=2.26in]{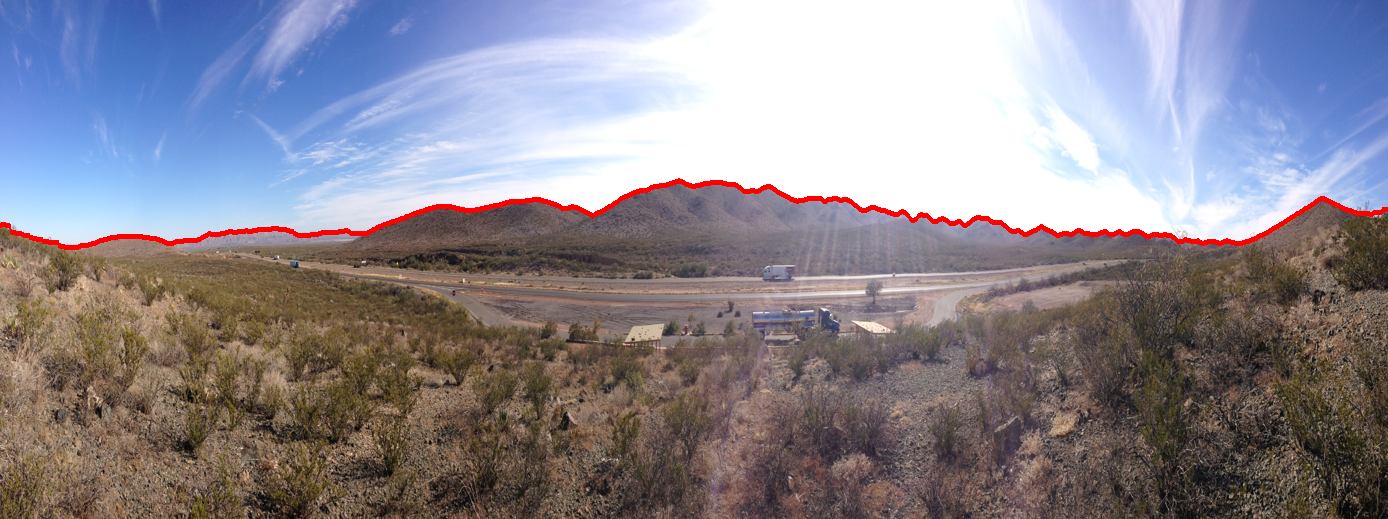}\\
(d) \hspace{2.22in} (e) \hspace{2.26in} (f)\\
%(c) \hspace{3.2in} (f)
\end{center}
\caption{Proposed Skyline Extraction Approach: (a) input image, (b) output of the Canny edge detector, (c) predicted score for each pixel belonging to skyline using selected linear filter based on pixel's structure tensor (brighter red and blue intensities respectively reflect more and less confidence for pixel belonging to the skyline), (d) gradient strength estimated as part of the structure tensor, (e) weighted predicted skyline score combined with gradient strength, (f) found skyline by dynamic programming overlaid on the original query image in red.}
\label{fig_proposed}
\end{figure*}

%------------------------------------------------------------------------
\section{Skyline Detection using Dynamic Programming}
\label{sec_2}
In this section, we outline the details of earlier skyline detection methods which address the underlying problem as graph-search and rely on dynamic programming to solve it. Since our proposed approach is also based on dynamic programming, it is important to describe the details of these methods from comparison view-point and building the basis of the proposed approach.  

\subsection{Skyline Detection as Graph Search Problem}
Lie \textit{et al.} \cite{Lie05} are the first one to formulate skyline detection as a multi-stage graph search problem. Given an $M \times N$ image, the image is first passed through an edge detector to compute a binary edge map $E(i,j)$ where $1/0$ imply the presence/absence of an edge pixel: 
\begin{equation}
E(i,j)=\begin{cases}
1,	&\text{if $(i,j)$ is an edge pixel},\\
0,	&\text{otherwise}.
\end{cases}
\end{equation}

They then used the edge map $E(i,j)$ to build an $M \times N$ multi-stage graph $G(V,E,\Psi,\Phi)$ where each pixel in the map corresponds to a graph vertex $v_{ij}$; a low cost $l$ is associated with edge pixels (vertices) while a very high cost (i.e., $\infty$) is associated with non-edge pixels as shown below:
\begin{equation}
\label{eq_nodal_cost}
\Psi(i,j)=\begin{cases}
l,	&\text{if $E(i,j)=1$},\\
\infty,&\text{otherwise}.
\end{cases}
\end{equation}

$\Psi(i,j)$ is the cost associated with vertex $i$ in stage $j$  (i.e., $v_{ij} \in V$). It should be noted that use of $\infty$ reflects a node with a high numeric cost. They have further used a gap filling process to address edge-gaps. Given a node $i$ in stage $j$, its neighborhood in the next stage $j+1$ is defined by a $\delta$ parameter, that is, the number of nodes to which $i$ could be connected in stage $j+1$. The edges from $i$ to its neighbors are associated with costs equal to the vertical absolute distance from it as shown in the equation below:
\begin{equation}
\label{eq_link_cost}
\Phi(i,k,j)=\begin{cases}
|i-k|, & \text{if $E(i,j)=E(k,j+1)=1$}\\
         & \text{and $|i-k| \leq \delta$},\\
\infty, & \text{otherwise.}
\end{cases}
\end{equation}

If a node $i$ in stage $j$ cannot be connected to any node in stage $j+1$ within $\delta$ distance, a search window is defined using two parameters: $\delta$ and tolerance-of-gap (tog). If an edge node $k$ is found within the search window, gap filling is performed by introducing dummy nodes between node $i$ in stage $j$ and node $k$ within the search window $j+$tog and a high cost is associated with such dummy nodes. More details for Lie \textit{et al.} can be found in \cite{Lie05} or \cite{Ahmad2020}. 

\subsection{Skyline Detection using Supervised Learning and Edges}
In a series of work \cite{Ahmad2013,Ahmad2014,Ahmad2015,Ahmad2015b,Ahmad2015c,Ahmad2017,Ahmad2020}, Ahmad \textit{et al.} extended the skyline detection approach of \cite{Lie05} by incorporating supervised machine learning techniques. Interested readers are requested to please consult their original papers \cite{Ahmad2013,Ahmad2014,Ahmad2015,Ahmad2015b,Ahmad2015c,Ahmad2017,Ahmad2020}.\par

In a simple extension of the binary edge map approach of Lie \textit{et al.}, to ensure good continuity, Ahmad \textit{et al.} \cite{Ahmad2014,Ahmad2015b} used the gradient information. They enforced the constraint that the difference between gradient magnitudes of adjacent pixels is minimized. The gradient magnitude at each pixel of the input image $I(i,j)$ is computed as follows:
\begin{equation}
\nabla(i,j)=\Gamma[I(i,j)], 
\end{equation}
where $\Gamma$ is a function which takes a gray scale image $I$ as input and returns the gradient magnitude image $\nabla$. Next, the difference of the gradient magnitude image $d\nabla(i,j)$ is computed:  
\begin{equation}
d\nabla(i,j)=|\nabla(i,j)-\nabla(i,j+1)|.
\end{equation}

The normalized (i.e., between 0 and 1) gradient magnitude and gradient difference images are combined through a weighted average:
\begin{equation}
G_r(i,j)=w_1*d\nabla(i,j)+(1-w_1)*(1-\nabla(i,j)),
\end{equation}
where $w_1$ is the weight parameter. Then they used the weighted average $G_r$ as the nodal cost:
\begin{equation}
\Psi(i,j)=G_r(i,j),
\end{equation}
whereas, the link costs may be initialized using equation (\ref{eq_link_cost}). In \cite{Ahmad2013,Ahmad2015b}, Ahmad \textit{et al.} considered Maximally Stable Extremal Edgess (MSEEs) and classification as a way to filter out non-skyline edges. They first applied Canny edge detector using various thresholds and kept only stable edges over a range of thresholds which they called MSEE edges. The MSEE edge map $E_m(i,j)$ is further filtered by classifying each MSEE pixel $(i,j)$ as belonging/not-belonging to the skyline using the trained classifier:
\begin{equation}
C(i,j)=\begin{cases}
1,	&\text{if $(i,j)$ pixel is classified as skyline},\\
0,	&\text{otherwise}.
\end{cases}
\end{equation}

In \cite{Ahmad2013,Ahmad2014,Ahmad2015b}, authors experimented with various texture features and their combinations for classification and found the SIFT-HOG combination yielded the lowest false negative rate. The edge map $E_+(i,j)$, comprising of the horizon classified MSEE edges, was used to define the nodal costs in the context of DP. Specifically, the edge map comprising of positively classified MSEE edge points can be expressed as follows:
\begin{equation}
  E_+(i,j)=\begin{cases}
    1, & \text{if $E_m(i,j)=1$ and $C(i,j)=1$},\\
    0, & \text{otherwise}.
  \end{cases}
\end{equation}

Authors have incorporated the classification information into nodal costs in two ways: (i) by using the binary costs based on $E_+(i,j)$ and (ii) by using the normalized classification scores $S(i,j)$ for these edges. For each case, the nodal-cost (\ref{eq_nodal_cost}) changes accordingly: 
\begin{equation}
\Psi(i,j)=\begin{cases}
l, & \text{if $E_+(i,j)=1$},\\
\infty, & \text{otherwise}.
\end{cases}
\end{equation}
\begin{equation}
\label{eq_classification_edge}
\Psi(i,j)=\begin{cases}
S(i,j), & \text{if $E_+(i,j)=1$},\\
\infty, & \text{otherwise},
\end{cases}
\end{equation}
whereas, by using $E_+(i,j)$ the link-cost (\ref{eq_link_cost}) adapts accordingly:   
\begin{equation}
\Phi(i,k,j)=\begin{cases}
|i-k|, & \text{if $E_+(i,j)=E_+(k,j+1)=1$}\\
         & \text{and $|i-k| \leq \delta$},\\
\infty, & \text{otherwise.}
\end{cases}
\end{equation}

They further combined classification scores with gradient information and adjusted the nodal-cost as below:
\begin{equation}
\Psi(i,j)=w_2*S(i,j)+(1-w_2)*G_r(i,j),
\end{equation}
where $w_2$ is a weight parameter.

\subsection{Skyline Detection using Dense Classification Score Images}
In the second set of their work \cite{Ahmad2015,Ahmad2015c,Ahmad2020}, Ahmad \textit{et al.} investigated to exclude the edge-detection and generated a dense classification score image (DCSI) which reflects the likelihood of a pixel belonging to the skyline. The resultant DCSI is then used directly to initialize the nodal-cost:
\begin{equation}
\label{eq_classification_edge_less}
\Psi(i,j)=S(i,j).
\end{equation}

The difference between Eqs. (\ref{eq_classification_edge}) and (\ref{eq_classification_edge_less}) is that in the former, classification score is used to initialize only those nodes which are MSEE edges and have been positively classified as skyline edges whereas in the latter, all the nodes have been initialized with the normalized classification scores without using any edge information. In the former case SIFT-HOG features were extracted around $(i,j)$ location while in the latter normalized pixel intensities were used.\par

For this work, they have investigated both Support Vector Machine (SVM) and Convolutional Neural Network (CNN) classifiers and resultant dense classification score images are identified by SVM-DCSI and CNN-DCSI respectively. For computational improvement, they further experimented by retaining only the best m-scores per stage (column) of the multi-stage graph which they termed as SVM-mDCSI and CNN-mDCSI. In extended work \cite{Ahmad2020} they investigated the fusion strategies by combining MSEE edges, Canny edges or gradient information with their DCSI images.

%%%%%%%%%%%%%%%%%%%%%%%%%%%%%%%%%%%%%%%%%%%%%%%%%%%%%%%%%%%%%%%%%%%%%%%%%%%%%%%%%%%%%%%%%%%%%%%%%%%%%%%%%%%%%%%%%%%%%%%%
%%%%%%%%%%%%%%%%%%%%%%%%%%%%%%%%%%%%%%%%%%%%%%%%%%%%%%%%%%%%%%%%%%%%%%%%%%%%%%%%%%%%%%%%%%%%%%%%%%%%%%%%%%%%%%%%%%%%%%%%

\begin{table*}
\centering \caption{Average Absolute Error for Existing and Proposed Approaches}
\begin{tabular}{|c|c|c|c|c|c|c|c|c|}
\hline
\textbf{Approach} & \multicolumn{4}{c|}{\textbf{Basalt Hills}} & \multicolumn{4}{c|}{\textbf{Web}}	\\  \cline{2-9}
 &  $\mu$ & $\sigma$  & $min$  & $max$ & $\mu$ & $\sigma$ & $min$  & $max$	\\  \cline{1-9}
 {\textbf{Edges Only} \cite{Lie05}} 			&	$5.55$	&	$9.46$	&	$0.53$	&	$49.31$	&	$9.15$ 	&	$17.92$	&	$0.38$	&	$93.02$\\  \hline
{\textbf{$G_r$} (\cite{Ahmad2014,Ahmad2015b})} 	&	$3.99$	&	$6.35$	&	$0.18$	&	$31.33$	&	$11.86$ 	&	$26.81$ 	&	$0.15$	&	$121.48$\\ \hline
{\textbf{SIFT+HOG Edges} (\cite{Ahmad2014,Ahmad2015b})} 			&	$0.57$	&	$1.02$	&	$0.00$		&	$3.58$	&	$0.87$ 	&	$1.03$ 	&	$0.43$	&	$7.05$\\ \hline
{\textbf{SIFT+HOG Scores} (\cite{Ahmad2014,Ahmad2015b})} 			&	$0.41$	&	$0.81$	&	$0.00$		&	$3.08$	&	$0.97$ 	&	$1.57$ 	&	$0.38$	&	$12.19$\\ \hline
{\textbf{SIFT+HOG Scores + $G_r$}  (\cite{Ahmad2014,Ahmad2015b})} 		&	$0.43$	&	$0.81$	&	$0.00$		&	$3.08$	&	$1.30$ 	&	$3.98$ 	&	$0.38$	&	$34.95$\\ \hline
{\textbf{SVM-mDCSI} (\cite{Ahmad2015,Ahmad2020})} 								&	$1.01$	&	$0.29$	&	$0.62$	&	$1.76$	&	$1.28$ 	&	$1.20$ 	&	$0.37$	&	$6.21$\\  \hline
{\textbf{CNN-mDCSI} (\cite{Ahmad2015,Ahmad2020})} 								&	$0.75$	&	$0.23$	&	$0.42$	&	$1.28$	&	$1.41$ 	&	$1.49$ 	&	$0.27$	&	$10.79$\\ \hline
{\textbf{SVM-DCSI+$G_r$} (\cite{Ahmad2020})} 					&	$0.60$	&	$0.29$	&	$0.17$	&	$1.31$	&	$4.86$ 	&	$15.98$ 	&	$0.14$	&	$98.46$\\ \hline
{\textbf{SVM-DCSI+MSEE Edges} (\cite{Ahmad2020})} 				&	$0.73$	&	$0.32$	&	$0.48$	&	$2.07$	&	$0.85$ 	&	$0.89$ 	&	$0.35$	&	$5.05$\\ \hline
{\textbf{SVM-DCSI+Canny Edges} (\cite{Ahmad2020})} 				&	$0.77$	&	$0.35$	&	$0.48$	&	$2.07$	&	$0.78$ 	&	$0.76$ 	&	$0.35$	&	$4.84$\\ \hline 
{\textbf{Proposed}} 	&	$1.45$	&	$1.55$	&	$0.24$	&	$5.69$	&	$7.85$ 	&	$32.64$ 	&	$0.29$	&	$203.34$\\
\hline
\end{tabular}
\label{tab_table1}
\end{table*}

%------------------------------------------------------------------------

\section{Shallow Learning}
\label{sec_3}
Our shallow learning is inspired and based on the BLADE framework introduced by Getreuer \textit{et al.} \cite{Getreuer2018} which is a generalization of popular super-resolution approach RAISR \cite{Romano2017}. BLADE has been demonstrated to perform well for various computational photography applications including denoising \cite{Choi2018}, demosaicing and image stylization \cite{Dorado2020}. Unlike earlier applications, we adopt this shallow learning framework for classification and explore its applicability for mountainous skyline extraction problem. Below, we briefly describe the inference and training steps of BLADE for a sample image restoration problem i.e., denoising and then extend these steps for our skyline extraction problem. 

\subsection{Spatially Varying Inference}
Let $\mathbf{z}$ and $\mathbf{u}$ be input (i.e., noisy image) and output (i.e., denoised image) gray-scale images respectively. A pixel at \textit{i}-th spatial location is specified by $\mathbf{z}_{i}$. Let there be a set of \textit{K} linear finite impulse response (FIR) filters denoted by $\mathbf{h^1}, \cdots,  \mathbf{h^k}$. The coefficients of these filters are learned during training phase. At inference time, a spatially varying filter of size $n \times n$ is  selected from the learned filter bank and applied on patch extracted around the central pixel $\mathbf{z}_{i}$. In vector notation, the inference can be written as a dot-product, 
\begin{equation}
\label{eq_1}
\hat{u}_i = (\mathbf{h}^{s(i)})^T\mathbf{R}_i\mathbf{z}, 
\end{equation}
where, $\mathbf{R}_i$ extracts a patch centered around pixel $i$ rearranged as vector, $\mathbf{h}^{s(i)}$ is the filter selected from the filter bank for the \textit{i}-th pixel based on its structure tensor, $(.)^T$ denotes the vector transpose operator and $\hat{u}_i$ is the inferred pixel value.\par

For computational photography and stylization tasks such as studied in \cite{Choi2018, Dorado2020, Getreuer2018}, the above operation is conducted for each pixel in the input image. For the skyline extraction problem, we first perform edge detection on the input image and then extract the gray-scale patches only around pixels detected as edges. The inferred pixel value $\hat{u}_i$ reflects the score of \textit{i}-th pixel belonging to the skyline.

\begin{figure*}[htbp]
\begin{center}
\includegraphics[width=2.65in]{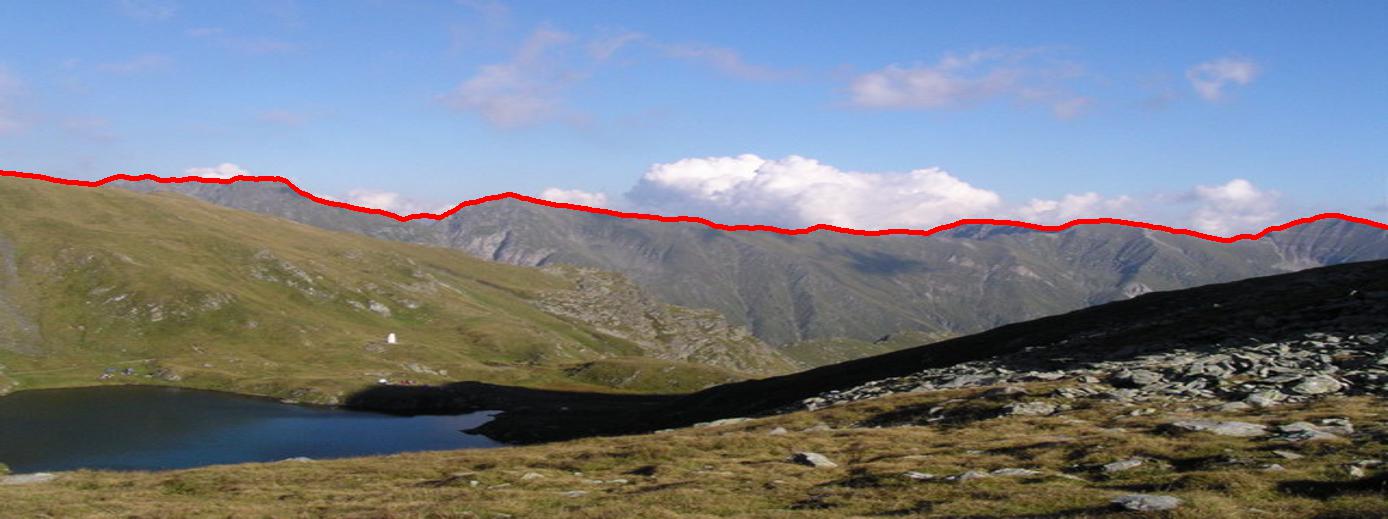}
\includegraphics[width=2.65in]{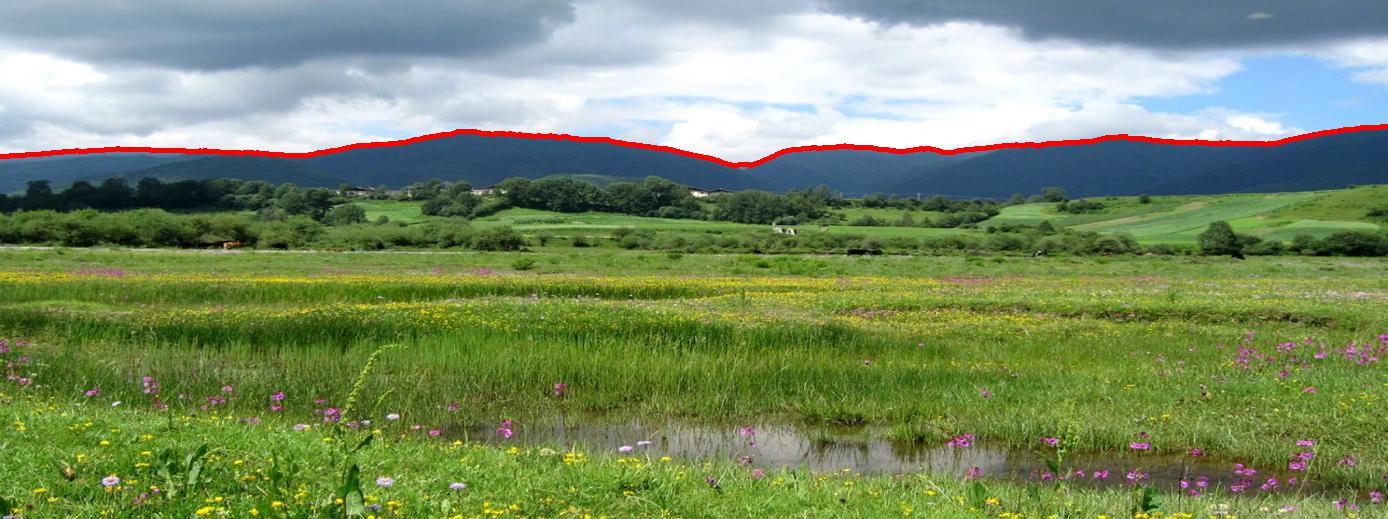}\\
\includegraphics[width=2.65in]{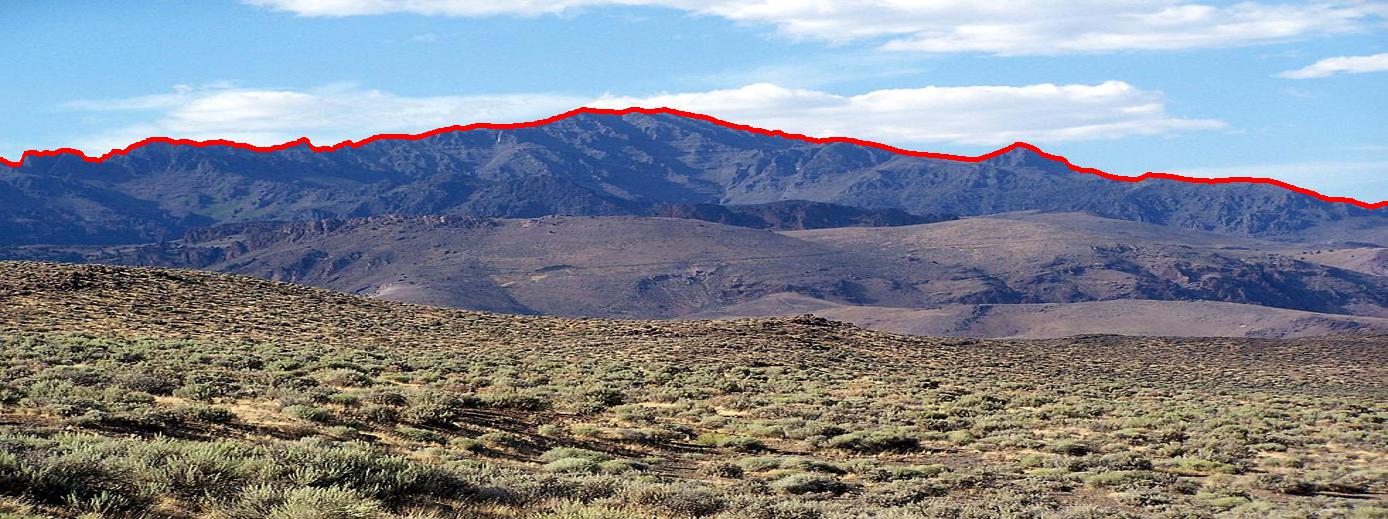}
\includegraphics[width=2.65in]{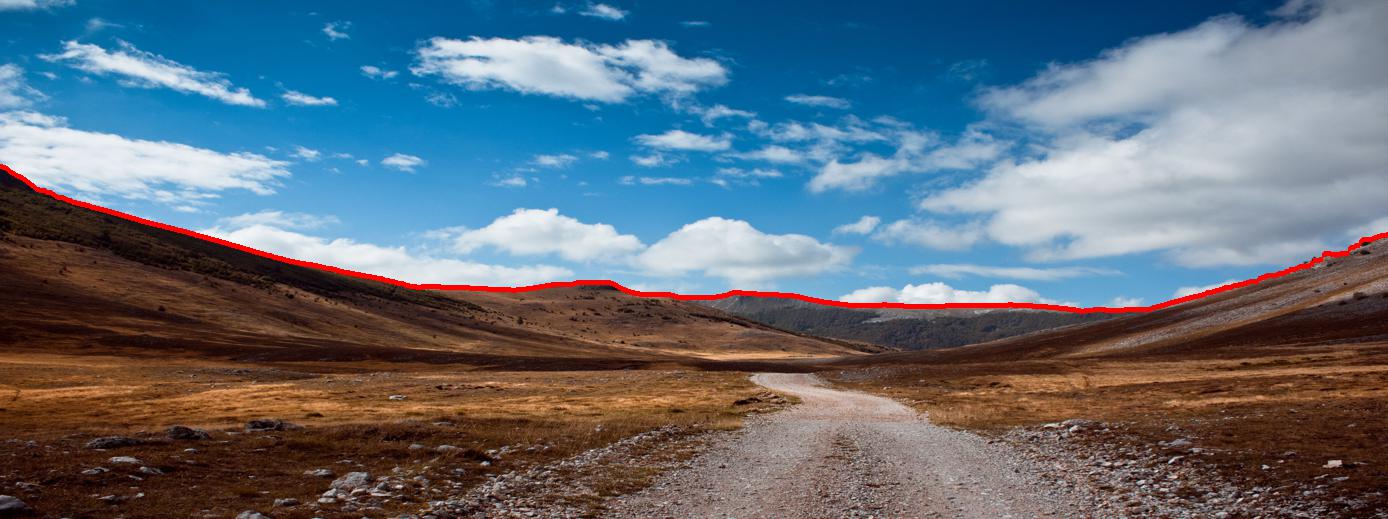}\\
\includegraphics[width=2.65in]{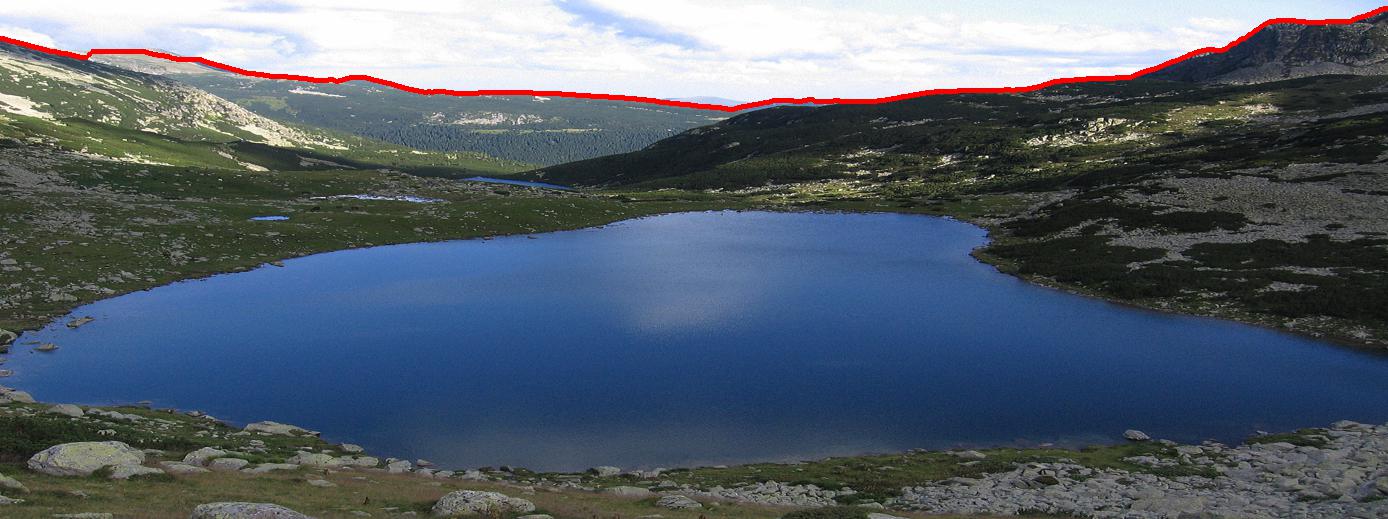}
\includegraphics[width=2.65in]{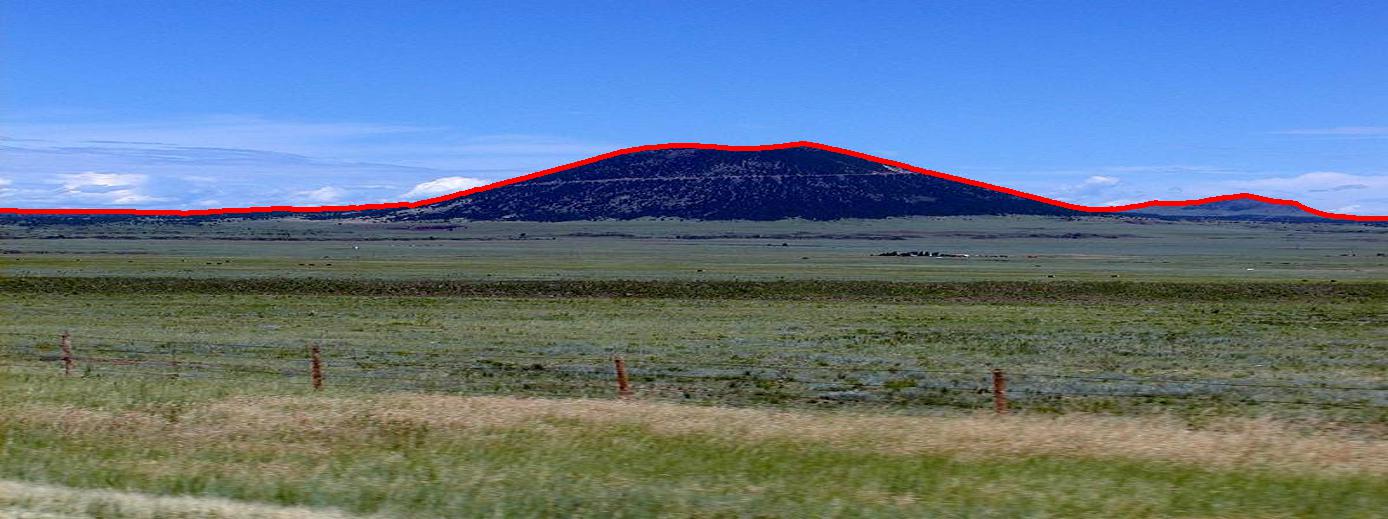}\\
\includegraphics[width=2.65in]{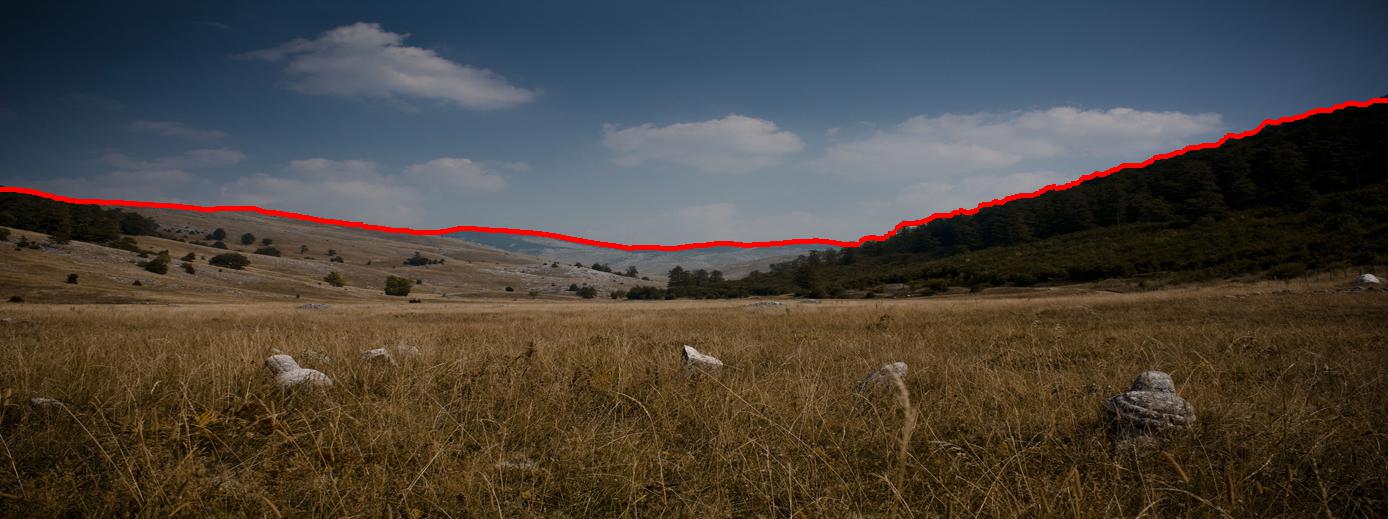}
\includegraphics[width=2.65in]{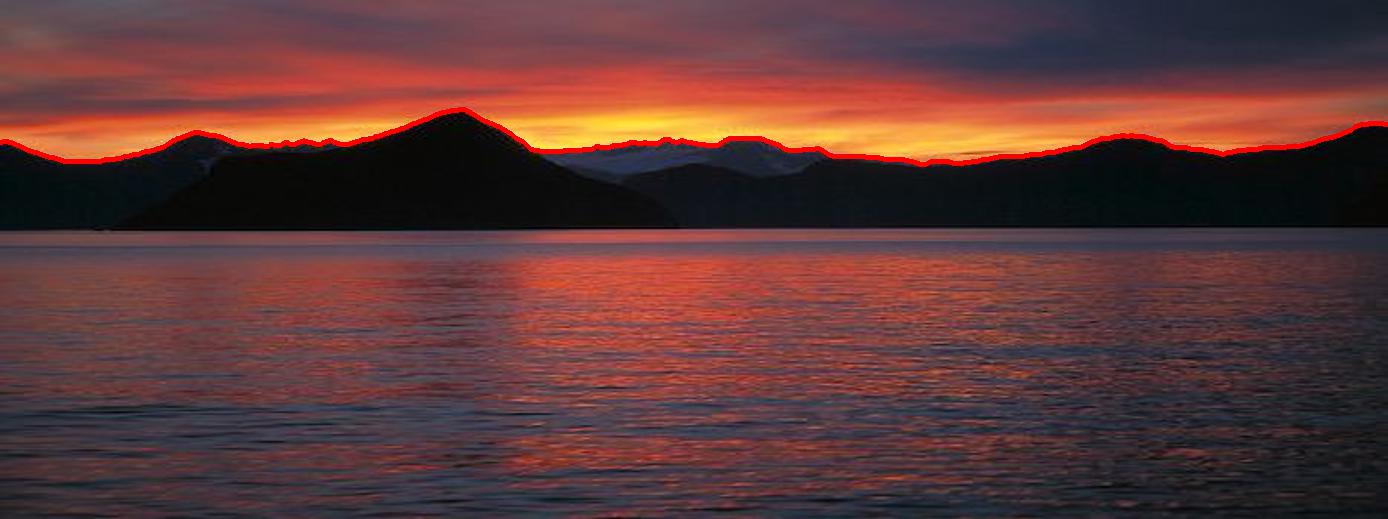}\\
\end{center}
\caption{Examples of good skyline detections by the proposed approach. Detected skylines are overlaid in red.}
\label{fig_good_detection}
\end{figure*}

\subsection{Learning the Filter Bank}
To learn the set of FIR filters $\mathbf{h^1}, \cdots,  \mathbf{h^k}$, training set comprising of input image and output target pairs is employed. Depending on the underlying application, $\mathbf{z}$ and $\mathbf{u}$ are the appropriate image pairs e.g., for learned denoising, $\mathbf{z}$ and $\mathbf{u}$ would be noisy and noise-free images respectively. The filter set is learned by minimizing the objective function comprised of $L^2$ loss and quadratic regularization term i.e.,       
\begin{equation}
\label{eq_2}
\argmin_{h^1, \cdots, {h^k}} \mathbf{\sum}_{k=1}^{K}(\mathbf{h}^k)^T\mathbf{Qh}^k + \mathbf{\sum}_{s(i)=k}(u_i -(\mathbf{h}^{k})^T\mathbf{R}_i\mathbf{z})^2, 
\end{equation}
where, $\mathbf{Q}$ determines the regularization and encourages the learned filters to be smooth. The minimization described in (\ref{eq_2}) can be solved independently for each filter and amounts to a multivariate linear regression with regularization listed below:
\begin{equation}
\label{eq_3}
\mathbf{h} = (\mathbf{Q + A}^T\mathbf{A})^{-1}\mathbf{A}^T\mathbf{b}.
\end{equation}

All the training examples belonging to a specific filter are accumulated in an $\mathbf{(n^2 + 1) \times (n^2 + 1)}$ gram matrix $\mathbf{G}$,
\begin{equation}
\label{eq_4}
\mathbf{G} \gets \mathbf{G} + \left(\begin{array}{c} \mathbf{R}_i\mathbf{z} \\ u_i \end{array}\right)
\left(\begin{array}{c} \mathbf{R}_i\mathbf{z}^T u_i \end{array}\right).
\end{equation}

Once the training samples are accumulated in $\mathbf{G}$, the respective matrices $\mathbf{A}^T\mathbf{A}$ and $\mathbf{A}^T\mathbf{b}$ for (\ref{eq_3}) can be accessed: 
\begin{equation}
\label{eq_5}
\left(\begin{array}{cc} \mathbf{A}^T\mathbf{A} & \mathbf{A}^T\mathbf{b}\\ \mathbf{b}^T\mathbf{A} & \mathbf{b}^T\mathbf{b} \end{array}\right) \gets  \mathbf{G}. 
\end{equation}

To learn the filters for skyline detection, we use positive training examples extracted along the ground truth skyline boundary and equal number of negative training examples extracted from random edge locations not in the vicinity of the skyline. We use two specific intensity levels as target $u_i$ for positive and negative examples. At inference time, after normalization, $\hat{u}_i$ reflects the confidence score of an edge-pixel belonging or not belonging to the skyline.      

\subsection{Filter Selection using Structure Tensor}
Following BLADE \cite{Getreuer2018}, we use local structure tensor to select the filter for a given pixel both during training and inference. Structure tensor is a way to provide local gradient estimate by employing Principal Component Analysis (PCA) of the gradients of \textit{i}-th pixel's local neighborhood $P_i$ by minimizing:
\begin{equation}
\label{eq_6}
\argmin_a \sum_{j\in_{P_i}}w^i_j(\mathbf{a}^T\mathbf{g}_j)^2, 
\end{equation}
where $\mathbf{g}_j$ is the gradient at \textit{j}-th pixel and $w^i_j$ is the spatial weighting e.g., coefficients of a Gaussian kernel. Instead of computing structure tensor using only Luma channel as done in \cite{Romano2017}, we adopt computing it jointly using all color channels as in \cite{Getreuer2018}. Using a $2 \times 3$ Jacobian matrix of color-wise gradients, 
\begin{equation}
\label{eq_7}
\mathbf{G}_j = \left[\begin{array}{ccc} \mathbf{g}_j^R & \mathbf{g}_j^G & \mathbf{g}_j^B \end{array}\right], 
\end{equation}
the unit vector $\mathbf{a}$ is found by minimizing (\ref{eq_6}) and replacing $\mathbf{g}_j$ with $\mathbf{G}_j$:
\begin{equation}
\label{eq_8}
\sum_{j}w^i_j\| \mathbf{a}^T\mathbf{G}_j \|^2 = \mathbf{a}^T (\sum_{j}w^i_j\mathbf{G}_j \mathbf{G}_j^T )\mathbf{a} = \mathbf{a}^T \mathbf{T}_i \mathbf{a}.
\end{equation}

The spatially weighted structure tensor $\mathbf{T}_i$ for \textit{i}-th pixel can be written as:  
\begin{equation}
\label{eq_9}
\mathbf{T}_i = \sum_{c} \sum_{j} w^i_j \left[\begin{array}{cc} g_{x,j}^c g_{x,j}^c & g_{x,j}^c g_{j,j}^c \\ g_{x,j}^c g_{y,j}^c & g_{y,j}^c g_{y,j}^c \end{array}\right], 
\end{equation}
where $c \in \left[R, G, B\right]$ the three color channels and $\left[g_{x,j}^c, g_{j,j}^c\right]^T = \mathbf{g}_j^c$. The three features computed from the eigen analysis of this $2 \times 2$ matrix $\mathbf{T}_i$ serve as the indices for feature selection after their respective quantization into pre-specified bins. The three features being computed are: (i) \textit{orientation} as the angle of the dominant eigenvector reflecting orientation of the gradient, (ii) \textit{strength} as the square root of the dominant eigenvalue $\sqrt{\lambda_1}$ reflecting gradient magnitude and (iii) \textit{coherence} characterizing the amount of anisotropy of local structure defined as:
\begin{equation}
\label{eq_10}
\frac{\sqrt{\lambda_1} - \sqrt{\lambda_2}}{\sqrt{\lambda_1} + \sqrt{\lambda_2}}, 
\end{equation}
where $\lambda_1 \geq \lambda_2 \geq 0$ are the eigenvalues of $\mathbf{T}_i$.

\begin{figure*}[ht]
\begin{center}
\includegraphics[width=2.26in]{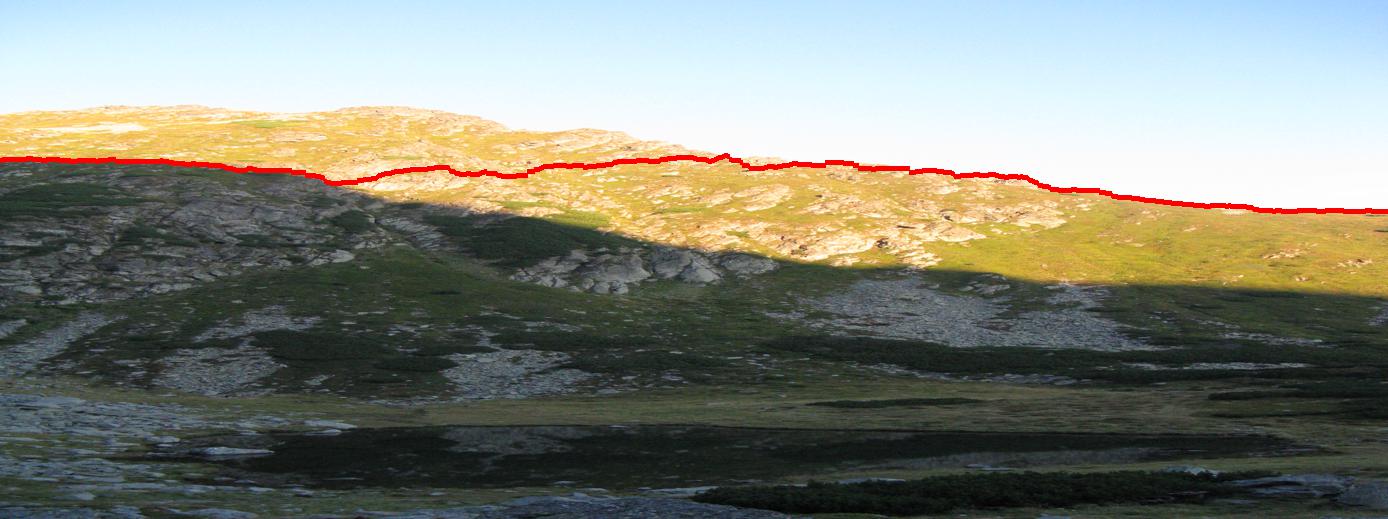}
\includegraphics[width=2.26in]{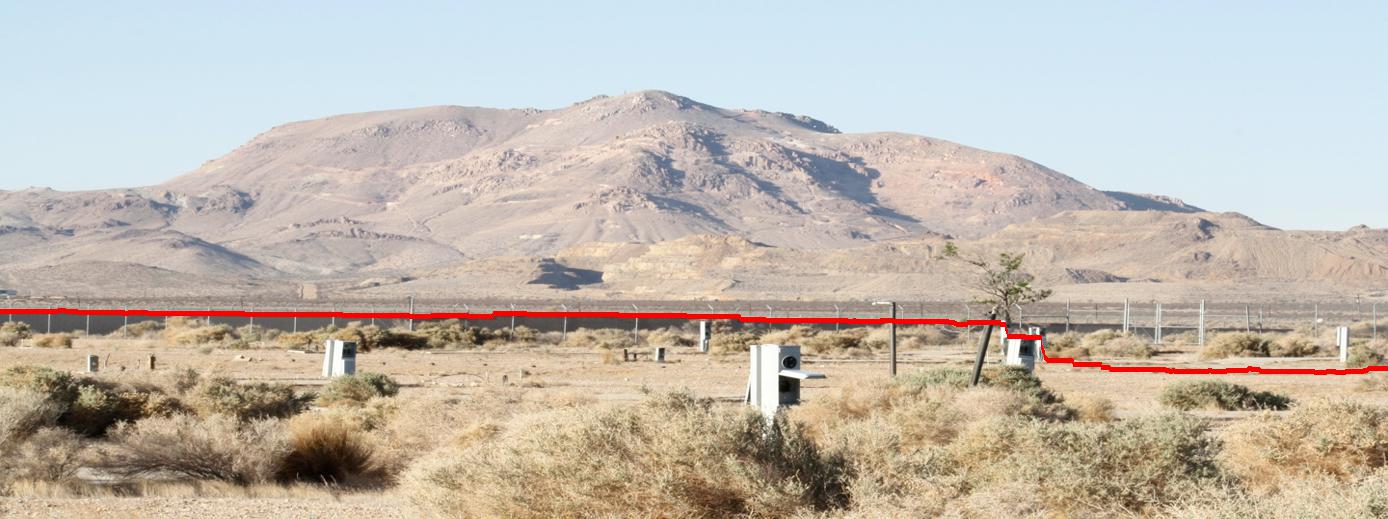}
\includegraphics[width=2.26in]{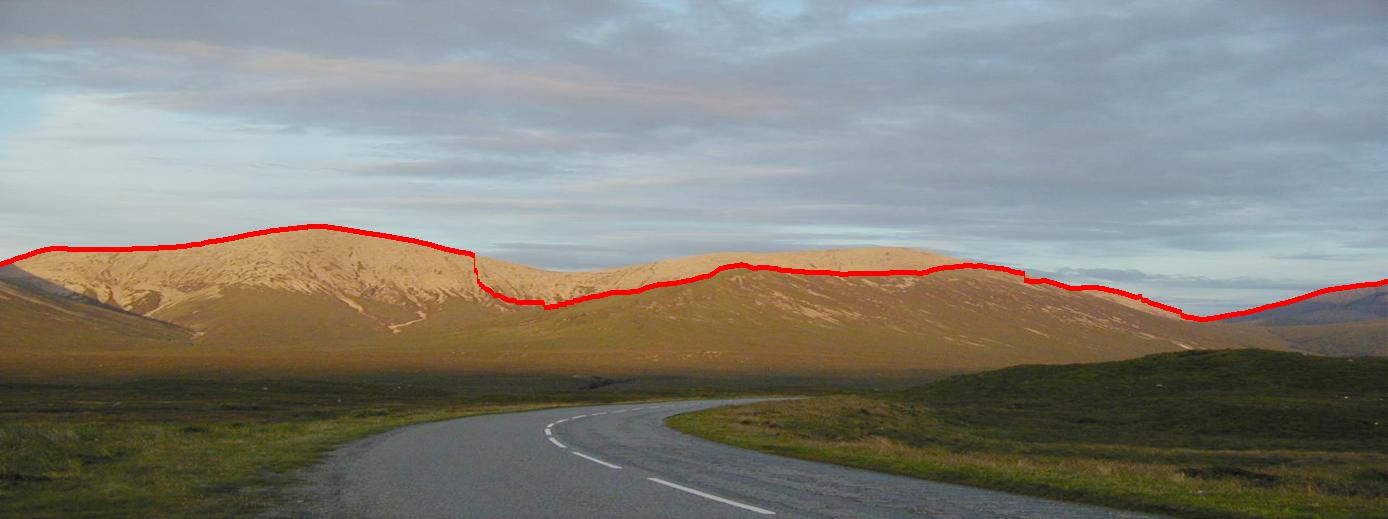}\\
\end{center}
\caption{Examples of faulty skyline detections by the proposed approach. Detected skylines are overlaid in red.}
\label{fig_bad_detection}
\end{figure*}

\begin{figure}[ht]
\centering
%\includegraphics[width=3.0in]{Basalt_data_set_performance.png}
%\includegraphics[width=3.0in]{Web_data_set_performance.png}\\
%(a) \hspace{3.0in} (b) \\
\includegraphics[width=3.3in]{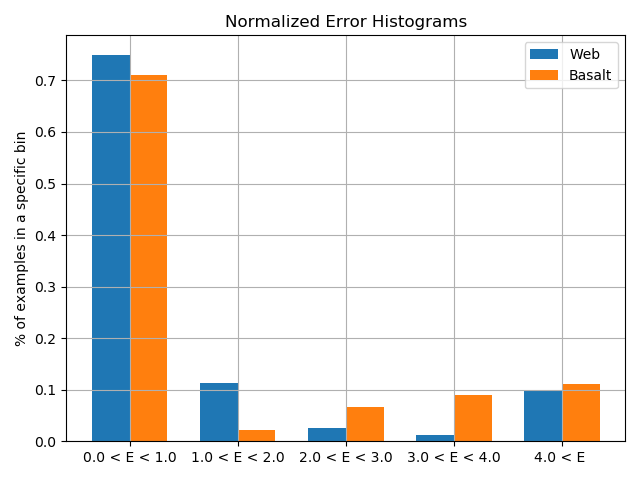}
\caption{Normalized histograms for Basalt Hills and Web data sets. In each case, for about $90\%$ of the images we get an average absolute pixel error of less than $4.0$ pixels which corresponds to a very good detection.}
\label{fig_normalized_histograms}
\end{figure}

\subsection{Proposed Skyline Detection Approach}
Unlike Ahmad \textit{et al.}, we do not rely on extracting feature descriptors (and their combinations) and then training SVM/CNN classifiers. Instead, for every given pixel, a linear filter is chosen from the learned filter bank based on its structure tensor and applied (dot product) to the patch around the central pixel. Further, we generate the prediction only for the pixels detected as edges by the edge-detector. In spirit, our approach is closer to that of SIFT-HOG classification score described in (\ref{eq_classification_edge}), however we use Canny edge detector instead of MSEE edges. Using Eqs. (\ref{eq_1}), (\ref{eq_nodal_cost}) and (\ref{eq_classification_edge}), the nodal-cost can be described as:
\begin{equation}
\label{eq_shallow_prediction_edge}
\Psi(i,j)=\begin{cases}
(\mathbf{h}^{s(k)})^T\mathbf{R}_k\mathbf{z}, & \text{if $E(i,j)=1$},\\
\infty, & \text{otherwise},
\end{cases}
\end{equation}
where, $k$ and $(i,j)$ correspond to the same central pixel in two different formulations. Although (\ref{eq_shallow_prediction_edge}) can be directly used to solve the shortest path problem through dynamic programming to find the skyline, we have found through experimentation that complementing the normalized prediction score with gradient strength improves the results. Since gradient magnitude $\sqrt{\lambda_1}$ is already computed as part of structure tensor analysis, we do not encounter any computational overhead and combine the normalized prediction score with the normalized gradient strength:
\begin{equation}
\Psi(i,j) = v * (1.0 - (\mathbf{h}^{s(k)})^T\mathbf{R}_k\mathbf{z})+ (1 - v) * (1.0 - \sqrt{\lambda_1(i,j)}), 
\end{equation}
where $v$ is a weight parameter. Both normalized gradient strength $[0, 1]$ and normalized prediction score $[0, 1]$ are subtracted from $1.0$ as we are trying to solve a minimization problem through dynamic programming. Figure \ref{fig_proposed} demonstrates the steps involved in our proposed skyline detection approach.

%------------------------------------------------------------------------

\begin{table*}[h]
\centering \caption{Average Absolute Error for Proposed Approach}
\begin{tabular}{|c|c|c|c|c|c|c|c|c|}
\hline
\textbf{Approach} & \multicolumn{4}{c|}{\textbf{Basalt Hills (Full)}} & \multicolumn{4}{c|}{\textbf{Web (73)}}	\\  \cline{2-9}
 &  $\mu$ & $\sigma$  & $min$  & $max$ & $\mu$ & $\sigma$ & $min$  & $max$	\\  \cline{1-9}
 {\textbf{Proposed}} 			&	$1.33$	&	$1.45$	&	$0.23$	&	$5.69$	&	$0.79$ 	&	$0.68$	&	$0.29$	&	$4.66$\\
\hline
\end{tabular}
\label{tab_table2}
\end{table*}

\begin{table*}[h]
\centering \caption{Impact of Filter Size on Skyline Detection Performance}
\begin{tabular}{|c|c|c|c|c|c|c|c|c|}
\hline
\textbf{Filter Size} & \multicolumn{4}{c|}{\textbf{Basalt Hills (Full)}} & \multicolumn{4}{c|}{\textbf{Web (73)}}	\\  \cline{2-9}
 &  $\mu$ & $\sigma$  & $min$  & $max$ & $\mu$ & $\sigma$ & $min$  & $max$	\\  \cline{1-9}
 {\textbf{$7 \times 7$}} & $1.3334$ & $1.4542$ & $0.23$ & $5.69$	& $0.7919$ & $0.6864$ & $0.29$	& $4.66$\\ \hline
 {\textbf{$9 \times 9$}} & $1.3402$ & $1.4559$ & $0.23$ & $5.69$	& $0.7921$ & $0.6610$ & $0.29$	& $4.33$\\ \hline
 {\textbf{$11 \times 11$}} & $1.3419$ & $1.4547$ & $0.23$ & $5.69$	& $0.7830$ & $0.6366$ & $0.29$	& $4.45$\\ \hline
 {\textbf{$13 \times 13$}} & $1.3557$ & $1.4639$ & $0.23$ & $5.69$	& $0.7887$ & $0.6691$ & $0.29$	& $4.54$\\ \hline
 {\textbf{$15 \times 15$}} & $1.3649$ & $1.4593$ & $0.23$ & $5.69$	& $0.7941$ & $0.6872$ & $0.29$	& $4.68$\\
\hline
\end{tabular}
\label{tab_table3}
\end{table*}

\section{Experimental Setup}
\label{sec_4}
\subsection{Data Sets for Skyline Detection}
There are several public data sets annotated for the problems of mountainous skyline detection and visual geo-localization which are briefly described below.   

\subsubsection*{Basalt Hills and Web}
The Basalt Hills data set is a subset of a larger data set captured by placing stereo cameras on an autonomous robot and navigating it through the Basalt Hills in California \cite{Nefian2014} to mimic a planetary-rover environment. Ahmad \textit{et al.} have manually generated ground truth for $45$ images from this data set and have used for their experiments. They have used $9$ out of these $45$ images as their training set and remaining $36$ along with the web set as their test set. The web data set is comprised of $80$ images gathered from the web for which Ahmad \textit{et al.} manually generated the ground truth skyline boundaries and used in their evaluations.  

\subsubsection*{CH1 and CH2}
CH1 \cite{Baatz2012} and CH2 \cite{Saurer2016} are visual geo-localization data sets which contain a total of about one thousand images with ground truth GPS location and FOV information. These datasets do not provide the ground truth camera orientation information and sky/non-sky segmentation is only available for $203$ images belonging to CH1 data set.    

\subsubsection*{GeoPose3K}
Brejcha and Čadík \cite{Brejcha2017} introduced the GeoPose3K data set that provides over three thousand images with ground truth GPS, orientation and semantic labels\footnote{Semantic labels are synthesized from a Digital Elevation Model.}.  
%As reported in paper, authors have manually annotated around $10\%$ of the  images while relied on developed algorithms to generate the ground truth for the rest. 
This is probably the biggest data set available for mountainous geo-localization. However, for skyline extraction, this is an extremely challenging data set as many images do not have visible skylines and it may be difficult for a human annotator to properly mark the ground truths for such images without any additional information e.g., relying on Digital Elevation Models (DEMs) or Google Earth. We feel this data set is suitable for camera orientation and GPS estimation where the visible portions of the terrain can be used to align with the DEMs, but it is challenging for skyline extraction approaches which have the underlying assumptions of skyline being visible and extending from left-to-right. The data set is also suited for evaluating general sky segmentation approaches as demonstrated in \cite{Ahmad2017}, where specifically designed learning-based skyline detection approaches performed poorly compared to scene parsing networks fine-tuned for sky segmentation.

\subsection{Evaluation Metric}
To quantitatively evaluate the performance of the proposed approach and compare against earlier approaches, we calculate the average pixel-wise absolute distance ($A_{err}$) between the detected and ground truth skylines:
\begin{equation}
A_{err}=\frac{1}{N}\sum_{j=1}^{N}{|P_{d(j)}-P_{g(j)}|}, 
\end{equation}
where $P_{d(j)}$ and $P_{g(j)}$ are the positions (rows) of the detected and ground truth skyline pixels in column $j$ and $N$ is the number of columns in the test image.\par

\begin{table*}[t]
\centering \caption{Impact of Training Set on Skyline Detection Performance}
\begin{tabular}{|c|c|c|c|c|c|c|}
\hline
\textbf{} & \multicolumn{6}{c|}{\textbf{Test Set}} \\ \cline{2-7}
\textbf{Training Set} & \multicolumn{2}{c|}{\textbf{CH1}} & \multicolumn{2}{c|}{\textbf{Basalt Hills (Full)}} & \multicolumn{2}{c|}{\textbf{Web}}	\\ \cline{2-7}
 &  $\mu$ & $\sigma$  & $\mu$ & $\sigma$  & $\mu$ & $\sigma$ \\ \cline{1-7}
 {\textbf{CH1}} & $-$ & $-$ &	$1.33$	&	$1.45$	& $7.85$ & $32.64$ \\  \hline
 {\textbf{Basalt Hills (Full)}} & $109.84$ & $144.34$ & $-$ & $-$ & $14.62$ & $54.81$ \\  \hline
 {\textbf{Web}} & $101.87$ & $140.18$ & $1.27$ & $1.39$ & $-$ & $-$ \\
\hline
\end{tabular}
\label{tab_table4}
\end{table*}

\subsection{Results}
\textbf{Base Results:} Unless otherwise stated, we use $203$ images from the CH1 \cite{Baatz2012} data set to learn the filter bank, where by default we used a filter size of $7 \times 7$ and $6$, $3$ and $16$ quantization bins for strength, coherence and orientation computed from structure tensor analysis. We use Web and Basalt Hills data set to report our base results. For Basalt Hills data set, we first report results for $36$ out of $45$ images to be consistent with other reported results for earlier methods and then separately report the results for all $45$ images. Table \ref{tab_table1} shows mean ($\mu$) and standard deviation ($\sigma$) of the evaluation metric for our proposed approach against other methods. For all of the compared methods, we take the reported numbers from the individual papers which are specifically noted along with the approach's name.\par 
As can be seen from table \ref{tab_table1}, our proposed approach performs poorer than other learning-based approaches where dedicated descriptors and non-linear classifiers are employed, whereas it outperforms other edge-based approaches which do not have any learning component. As reported in \cite{Getreuer2018,Choi2018}, the BLADE framework is well-suited for resource constrained platforms e.g., mobile devices. Inheriting from BLADE, our skyline extraction approach provides a good trade-off between performance and computation and is suitable for applications with limited on-line memory and compute resources e.g., augmented reality, visual geo-localization and navigation of UAVs and planetary rovers. To better understand the performance of our approach, we plot the normalized histograms of the average absolute error for Basalt and Web data sets in figure \ref{fig_normalized_histograms}. It is clear that for only $10\%$ of the images in each data set, we get an average pixel error of $4.0$ or more. Figure \ref{fig_good_detection} shows some of the good skyline detection results for the Web data set, whereas some failure cases of the proposed approach from the same data set are shown in figure \ref{fig_bad_detection}. As mentioned before, Ahmad \textit{et al.} reported results for $36$ out of $45$ images for Basalt Hills data set. For a direct comparison, in table \ref{tab_table1}, we also reported the error for our approach for the same $36$ images, however, for completeness table \ref{tab_table2} reports results for all $45$ images identified as \textbf{Basalt Hills (Full)}. For Web data set, our approach resulted into an error of $4.0$ pixels or more for $7$ out of $80$ images; excluding these images from the error computation results in mean and standard deviation which are directly comparable to the existing computationally intensive learning based methods. The error metric for these $73$ web images are also reported in table \ref{tab_table2} identified by the column \textbf{Web (73)}. We have identified number of reasons for partial or full failure cases. Firstly skyline or portion of skyline are not strong edges and are missed by the edge detector and subsequently a prediction score is not generated for such pixels. Secondly the employed training data is limited which caused some of the feature bins to be not learned very well. At test time if a pixel falls to such a bin based on its structure tensor, the resultant prediction score might not be accurate. Finally there exist a shortest path in the image extending from left-to-right which is less expensive than the skyline path and hence been found by dynamic programming as the solution (e.g, middle image in figure \ref{fig_bad_detection}). 

\textbf{Filter Size:} Since filter size is directly related to the computational overhead, following \cite{Getreuer2018,Choi2018} we have used a filter size of $7 \times 7$ for our base experiments. We conducted an experiment to evaluate the impact of filter size on skyline detection performance. Table \ref{tab_table3} shows the performance of skyline detection on \textbf{Basalt Hills (Full)} and \textbf{Web (73)} as the filter size is varied. As evident from table \ref{tab_table3}, there is no noticeable effect on skyline detection due to change in the filter size. \par 

\textbf{Data Set Variation:} We have demonstrated results where CH1 data set has been used as a training set and Basalt Hills and Web data sets have been used as the test sets. To study the affect of the training set, we conducted an experiment where each of the three data sets serves as a training set while the remaining two are used as the test sets. Table \ref{tab_table4} demonstrates the results for this experiment. As expected the performance on the test set is proportional to the versatility and number of images in the used training set, e.g., in case of Basalt as a training set, the performance of skyline detector on Web set is poorer than when CH1 data set is used as training set. Similarly, on CH1 data set the performance is relatively better when trained on Web set compared to when trained on Basalt set. The overall skyline detection performance on CH1 data set is poor compared to Basalt and Web data which is understandable as CH1 data set is more challenging compared to the other two.\par 

\textbf{GeoPose3K:} As mentioned earlier the GeoPose3K dataset \cite{Brejcha2017} is more suited for general scene parsing approaches where sky segmentation is the underlying objective instead of skyline detection. Nonetheless, we provide the comparison of our proposed approach on GeoPose3K in table \ref{tab:table5} where we report the numbers for existing approaches from \cite{Ahmad2017} (Table III in that paper). In \cite{Ahmad2017}, authors fine-tuned general scene-parsing deep networks for sky segmentation using CH1 dataset and provided a comparison between scene-parsing and specific skyline detection methods using 2895 images from the GeoPose3K dataset. Following \cite{Ahmad2017} we also report the performance of our proposed approach on the same 2895 images in table \ref{tab:table5} where we used CH1 dataset to learn the filer bank. In addition to the pixel-wise absolute distance, we also report the segmentation accuracy to be consistent with \cite{Ahmad2017}. As expected our approach is out-performed by scene parsing approaches adapted for sky segmentation while comparable to specifically designed skyline detection methods e.g., Horizon-DCSI-CH1.

\begin{table}
\centering \caption{Comparison of the proposed approach on GeoPose3K dataset}
\begin{tabular}{|c|c|c|c|}
\hline
\textbf{Approach} & {\textbf{Accuracy}} & \multicolumn{2}{c|}{\textbf{Absolute Error}}\\  \cline{3-4}
 &  & $\mu$  & $\sigma$  \\  \cline{1-4}
{\textbf{FCN8s-Pascal}\cite{Ahmad2017}} 								&	$0.9551$	&	$32.161$	&	$57.510$\\ \hline 
{\textbf{FCN16s-Pascal}\cite{Ahmad2017}} 							&	$0.9539$	&	$32.888$	&	$58.193$\\ \hline 
{\textbf{FCN32s-Pascal}\cite{Ahmad2017}} 							&	$0.9520$	&	$33.534$	&	$57.588$\\ \hline 
{\textbf{FCN8s-SiftFlow-g}\cite{Ahmad2017}} 							&	$0.9491$	&	$34.975$	&	$53.334$\\ \hline 
{\textbf{FCN8s-SiftFlow-s}\cite{Ahmad2017}} 							&	$0.9563$	&	$31.399$	&	$55.052$\\ \hline 
{\textbf{Horizon-ALE-CH1}\cite{Ahmad2017}} 							&	$0.9411$	&	$43.959$	&	$86.038$\\ \hline 
{\textbf{Horizon-DCSI-CH1}\cite{Ahmad2017}} 							&	$0.8743 $	&	$99.742$	&	$160.252$\\ \hline  
{\textbf{SegNet-CH1}\cite{Ahmad2017}} 								&	$0.8437$	&	$114.893$	&	$99.021$\\ \hline  
{\textbf{FCN8s-SiftFlow-s-CH1}\cite{Ahmad2017}} 						&	$0.9486$	&	$37.947$	&	$69.435$\\ \hline  
{\textbf{FCN8s-Pascal-CH1}\cite{Ahmad2017}}							&	$0.9432$	&	$41.596$	&	$71.707$\\ \hline
{\textbf{Proposed}}							&	$0.8652$	&	$105.712$	&	$164.224$\\

\hline
\end{tabular}
\label{tab:table5}
\end{table}

\subsection{Resource Comparison}

In table \ref{tab:table6}, we report the memory foot-print and the inference time for our proposed approach and compare it against selective existing methods. Specifically, we compare against top-performing deep networks (table \ref{tab:table5}) adapted for sky-segmentation \cite{Ahmad2017} and a representative approach designed for skyline detection i.e., Horizon-DCSI \cite{Ahmad2015}. For each approach we report an average inference time for an image size of $519 \times 1388$ pixels using a consistent single CPU environment (Processor: 1.8 GHz Intel Core i5, Memory: 8GB 1600 MHz DDR3). It should be noted that we do not include the time taken to load the coefficients/weights of a model. As clear from \ref{tab:table6}, each of the deep networks adapted for sky segmentation has a memory foot-print of more than 500 megabytes whereas the average inference time for specifically designed skyline detection approach is more than 20 seconds. Compared to these methods, our approach provides a middle ground where the memory requirement for our $7 \times 7$ filter bank is just 57KB and inference time is the best, collectively rendering our approach best suited for resource constrained mobile devices where memory, inference time and battery life are of great concern. We should note that both Horizon-DCSI and our proposed approach can further benefit from GPU implementation as there are several identical operations being performed for every pixel in the image.           

\begin{table}
\centering \caption{Comparison of required computational resources of the proposed approach against others}
\begin{tabular}{|c|c|c|}
\hline
\textbf{Approach} & {\textbf{Memory (MB)}} & {\textbf{Inference Time (s)}}\\  \hline
{\textbf{FCN8s-Pascal}} 	&	$513$	&	$15.39$\\ \hline
{\textbf{FCN16s-Pascal}} 	&	$514.3$	&	$15.31$\\ \hline
{\textbf{FCN32s-Pascal}} 	&	$519.4$	&	$16.16$\\ \hline
{\textbf{FCN8s-SiftFlow}} 	&	$514$	&	$15.65$\\ \hline
{\textbf{Horizon-DCSI}} 	&	$0.0022$	&	$20.45$\\ \hline
{\textbf{Proposed ($7 \times 7$)}} 	&	$0.057$	&	$10.59$\\ \hline
%{\textbf{Proposed ($9 \times 9$)}} 	&	$0.093$	&	$-$\\ \hline
%{\textbf{Proposed ($11 \times 11$)}} 	&	$0.140$	&	$-$\\ \hline
%{\textbf{Proposed ($13 \times 13$)}} 	&	$0.195$	&	$-$\\ \hline
%{\textbf{Proposed ($15 \times 15$)}} 	&	$0.259$	&	$-$\\ \hline

\end{tabular}
\label{tab:table6}
\end{table}
\section{Conclusions}
\label{sec_5}
Earlier skyline detection approaches are based on supervised or deep learning and are unsuitable for resource constrained devices. In this paper, we have presented a computationally efficient and faster skyline detection approach which is based on the shallow learning framework specifically designed for mobile and resource constrained devices. Our approach provides a good trade-off between performance and computations, as it outperforms non-learning skyline detection methods while comparable to supervised and deep learning based methods. We have provided a quantitative comparison of our proposed approach against earlier relevant skyline detection methods using four publicly available data sets and established performance metrics. We conducted experiments to study the affect of training data and filter size. Further we provided a resource comparison (in terms of memory foot-print and inference-time) of our approach against existing ones. In future work, we would investigate to improve the performance of our approach while maintaining the same computational foot-print. To this end, we would like to understand reasons behind the failure cases in addition to ones identified above. In our base experiments, the performance of the proposed approach on CH1 data set is rather poor which is related to less versatility of the training set (i.e., Web and Basalt data sets). This would be another dimension of our exploration where we aim to remedy the failure rate by generating ground truth for CH2 data set. Additionally we intend to isolate samples from GeoPose3K where humans can annotate skylines without additional information so that methods designed specifically for skyline detection can be correctly evaluated.      

\section*{Acknowledgements}
This work was supported by project no. LTAIZ19004 Deep-Learning Approach to
Topographical Image Analysis; by the Ministry of Education, Youth and Sports
of the Czech Republic within the activity INTER-EXCELENCE (LT), subactivity INTER-ACTION (LTA), ID: SMSM2019LTAIZ.
Computational resources were partly supplied by the project e-Infrastruktura CZ (e-INFRA LM2018140) provided within the program Projects of Large Research, Development and Innovations Infrastructures.

\bibliographystyle{ieeetr}
\bibliography{egbib}

\end{document}